\documentclass[runningheads]{llncs}
\usepackage[toc]{appendix} % For the Appendix
\usepackage{enumerate} % To create lists
\usepackage{graphicx}
\usepackage{amsmath}
\usepackage{multicol} % To split text into columns
\usepackage{multirow}
\usepackage{subcaption}
\usepackage{dirtytalk}
\usepackage{url}
\usepackage{float}
\usepackage{csquotes}
\usepackage{setspace}
\usepackage{pifont}% http://ctan.org/pkg/pifont
\newcommand{\cmark}{\ding{51}}%
\usepackage{adjustbox} % rotate table
\usepackage{comment} % package to comment multiple lines
\begin{document}
\emergencystretch 3em
\title{Using Natural Language Processing and Networks to Automate Structured Literature Reviews: An Application to Farmers Climate Change Adaptation}

\titlerunning{Using NLP and Networks for Literature Reviews}

\author{Sofia Gil-Clavel\inst{1}\orcidID{0000-0003-4707-849X} \and\\
Tatiana Filatova\inst{1}\orcidID{0000-0002-3546-6930}}
\authorrunning{S. Gil-Clavel and T. Filatova}
\institute{Faculty of Technology, Policy and Management, \\Delft University of Technology, The Netherlands\\
\email{\{B.S.GilClavel,T.Filatova\}@tudelft.nl}\\}
\maketitle              % typeset the header of the contribution
\begin{abstract}
The fast-growing number of research articles makes it problematic for scholars to keep track of the new findings related to their areas of expertise. Furthermore, linking knowledge across disciplines in rapidly developing fields becomes challenging for complex topics like climate change that demand interdisciplinary solutions. The rise of machine-learning-supported text analysis has been instrumental in processing thousands of articles. Yet, how text relationships are built remains a Black-Box for domain experts, making it difficult to relate connected concepts to existing theories conceptualizing cause-effect relationships and permitting hypothesis building and testing. This paper presents an approach to sensibly use Natural Language Processing by extracting variable relations and synthesizing their findings using networks while relating to key concepts dominant in relevant disciplines. As an example, we apply our methodology to analyze farmers’ adaptation to climate change and compare the results with mainstream text summarization methods. Furthermore, we validate our methodology by asking experts to score and compare the outcomes. Results show that using Natural Language Processing and networks descriptively offer an interpretable way to synthesize literature review findings that outperform mainstream text summarization methods. This methodology gives not only the direction and the frequency of the association between the words but also the frequency the word appears in the articles, which is instrumental when performing literature reviews.

\end{abstract}
\section{Introduction}

The fast expansion of research articles hinders researchers from keeping track of the emerging findings in their areas of expertise \cite{larsen_rate_2010}. While systematic literature reviews are labor-intensive, the median time for them to go outdated is 5.5 years \cite{shojania_how_2007}. Given the current pace of publishing, it is increasingly challenging to cover the escalating pool of articles and to monitor the latest developments in a particular field. Automated content analysis \cite{nunezmir_automated_2016} and extractive text summarization systems \cite{bui_extractive_2016} have been argued as key tools to support literature synthesis and to keep track of research trends.

Automated content analysis for literature reviews commonly aims to map text into topics using the articles’ abstracts. Nunez-Mir et al. \cite{nunezmir_automated_2016} show how automated content analysis can synthesize big volumes of literature’s abstracts from ecology and evolutionary biology. In the climate change domain, Creutzig et al. \cite{creutzig_reviewing_2021} use machine learning and experts evaluations to map key messages relevant to climate change mitigation using around 100,000 articles’ abstracts related to demand-side mitigation. Regarding climate change adaptation, Nalau et al. \cite{nalau_mapping_2021} use the abstracts and metadata from around 7000 articles to map the evolution of climate change adaptation science across time.

Text summarization systems are normally used for more in-depth literature reviews that use the full articles’ text. In this context, a summarization system takes a predefined set of documents and generates a summary, which can be extractive (the system extracts and ranks the sentences from the original text), abstractive (the system extracts knowledge from the original text and reconstructs them into a new piece), or hybrid (combines both approaches by rewriting a summary of the sentences chose with the extractive approach) \cite{koh_empirical_2023,wang_systematic_2021,batista_quantitative_2015}. Wang et al. \cite{wang_systematic_2021} offers a systematic review of the different methodologies for Electronic Health Records and biomedical literature summarizations. Based on their systematic review, 83\% of the articles analyzed used at least the full text as input. As summary output, 81\% of the articles adopted an extractive approach, while 17\% of the articles used an abstractive approach – the remaining 2\% compared both outputs. 
%\cite{koh_empirical_2023} \textbf{perform a survey of the summarization systems available.}

However, automated content analysis and extractive text summarization systems stem from machine learning and deep learning methodologies. These approaches are often referred to as a `Black-Box' because they make it difficult to theorize based on their results and to understand how text relationships are built \cite{bender_dangers_2021,rudin_stop_2019}. As they tend to rank and re-organize the text, they hinder interpretation by domain expert. On the one hand, this hinders the understanding of the role of the extracted piece in the sentence and possible causal relationships explained in the text \cite{koh_empirical_2023,wang_systematic_2021}. On the other hand, they might contain redundant information, which may result in a deficiency of information \cite{koh_empirical_2023,wang_systematic_2021,batista_quantitative_2015}. Furthermore, these methodologies can be opaque \cite{koh_empirical_2023,burrell_how_2016}, in terms of what text features the algorithm relied on to perform classifications. This leads to calls to stop using Black-Box methodologies in favor of interpretable algorithms \cite{rudin_stop_2019} grounded in theory and that use tailored databases, as opposed to trained over all the data found in the web \cite{bender_dangers_2021}.

In this work, we argue that using Natural Language Processing (NLP) descriptively (as opposed to predictively) offers interpretable methodologies to perform literature reviews when the results are sensibly visualized using networks. NLP is a range of computational techniques whose main goal is to provide machines with the capability to understand natural language by understanding its semantics. This means that “the machine understand[s] not just the statistical properties but also the meaning and context of, say, a certain word” \cite[p.2]{jain_introduction_2022}. At the same time, in its simplest form, a network is a collection of points (nodes) joined together in pairs by lines (edges). In research, many systems can be represented as networks, for example, the internet, social interactions, and collections of computers. Network analysis can uncover system patterns through their different structures, properties, and metrics \cite{newman_networks_2017}. 

The combination of NLP and networks to summarize text is not new. In their pioneering work, Shojania et al. \cite{shojania_how_2007} use NLP and networks to generate visual overviews of unstructured text from novels and blogs to investigate different research questions, such as “Who was involved in the debate about competing scientific ideas?” In more recent work, Sofoluwe et al. \cite{smith_natural_2021} use NLP and networks to provide insights on policy and scientific discourse around Sustainable Development Goals. In their network, they use NLP to identify topics in the text and then build the network using nodes to represent the topics and edge weights to represent the cosine similarity between the topics. However, to our knowledge, networks have not been used to generate visual overviews of articles' findings to perform literature reviews.

In this work, we propose a methodology based on NLP and network visualization to summarize articles' findings in a descriptive and interpretable manner. In the rest of the article, first, we introduce our NLP-supported literature review algorithm. Second, we demonstrate its utility by applying it to an articles' database about farmers' climate change adaptation. Third, we explain the logic behind the network visualization and use it to visualize the results from the NLP-supported literature review. Fourth, we compare and validate the algorithm against results from text summarization systems by asking climate change experts to score the outcomes. Lastly, we draw the conclusions and finalize with the limitations and future work.

\section{Methodology} 

This section introduces our NLP-supported literature review algorithm while applying it to the articles' database. The articles' database focuses on drivers of farmers' adaptation to climate change. In the literature, there are some reviews on what the agricultural farmers' adaptation measures are \cite{bahinipati_evidence-based_2021,below_micro-level_2010,shaffril_systematic_2018} and what motivates farmers to adapt \cite{dang_factors_2019}. Therefore, we apply our methodology to 43 of the 47 articles\footnote{We did not have access to or could not find four of the articles.} used by Dang et al. \cite{dang_factors_2019} for their Literature Review (Table \ref{tab:Tab1}, Appendix A). This way, we can use Dang's et al. \cite{dang_factors_2019}'s findings as our gold standard when comparing and validating the results from our methodology against summarization methods.

\subsection{General Overview of the Algorithm}

\begin{figure}
\centering
\includegraphics[width=0.6\textwidth]{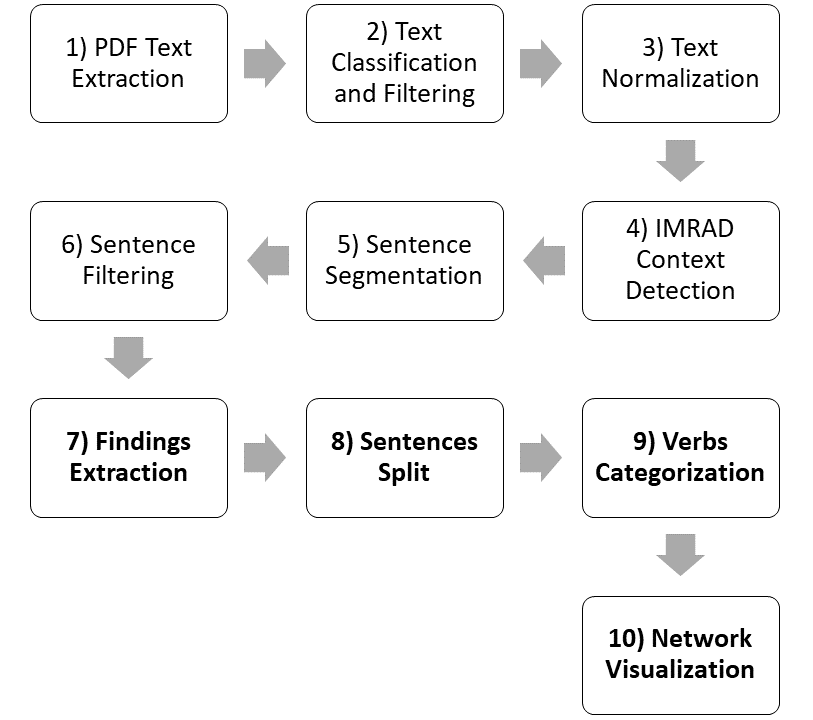}
\caption{Flow chart diagram presenting our algorithm based on Bui et al. \cite{bui_extractive_2016}. Boxes with bold text signal the novel steps added in the framework presented here.} \label{fig2}
\end{figure}

We build upon Bui et al. \cite{bui_extractive_2016} extractive text summarization system to perform the NLP-supported systematic literature review. In this work, we followed Bui's et al. \cite{bui_extractive_2016} first six steps and, at the end, introduced four new ones (Fig. \ref{fig2}): 1) PDF Text Extraction; 2) Text Classification and Filtering; 3) Text Normalization; 4) IMRAD Context Detection; 5) Sentence Segmentation; 6) Sentence Filtering; 7) Findings Extraction; 8) Sentences Split; 9) Verbs Categorization; and 10) Network Visualization. In the following paragraphs, we further explain each step from Fig. \ref{fig2} while using the data from our application as an example.

First, {\itshape PDF Text Extraction}, as the name hints, transforms the pdf file into plain text. For this, we used the Python package \texttt{pdfminer.six} \cite{guglielmetti_pdfminersix_2022}. The result of this step is a data frame where the rows are the articles, and the columns are the articles’ names and their cleaned plain text. Second, {\itshape Text Classification and Filtering} categorizes the extracted information into title, abstract, body text, semi-structure, and metadata. For this, we transformed the articles’ metadata (doi, title, abstract, authors, etc.) into a data frame where the rows are the articles and the columns are the articles’ metadata values by article. Then, we merged this with the cleaned plain text database from the previous step. This resulted in a single database with all the metadata and text by article.

Third, {\itshape Text Normalization} means transforming the text into canonical form. For this, we cleaned the data using regular expressions to remove: non-ASCII characters, URLs, and in-text references, such as “(NAME, YEAR)” and “et al. (YEAR).” We also used regular expressions to remove all the strings that referred to the journal's information and copyrights. Besides this, we used the Spacy component \texttt{AbbreviationDetector}, from the \texttt{scispacy} package \cite{intelligence_scispacy_nodate}, and the python package \texttt{neuralcoref} \cite{noauthor_neuralcoref_nodate}. \texttt{AbbreviationDetector} detects and expands all text abbreviations and \texttt{neuralcoref} substitutes any ambiguous expressions, such as pronouns, with the real-world entities they are referring to. Fourth, {\itshape IMRAD Context Detection} assigns text snippets into standard scientific articles' categories named IMRAD. IMRAD stands for Introduction, Methods, Results, and Discussion \cite{sollaci_introduction_2004}. For this, we implemented an algorithm to split the articles’ cleaned plain text into the IMRAD categories. Each of these sections was saved as a new independent column in the database.

Fifth, {\itshape Sentence Segmentation} involves tokenizing the sentences’ elements. Here, we used the python package \texttt{spaCy} \cite{spacy_spacy_2022} to tokenize the sentences and to extract the Parts-Of-Speech of each of the articles’ sections. Specifically, we used the \texttt{scispacy} model \texttt{en\_core\_sci\_lg} \cite{neumann_scispacy_2019}. {\itshape Sentence Filtering} involves filtering out irrelevant information. In our work, this translated into keeping the columns \textit{abstract}, \textit{findings}, \textit{conclusions}, and \textit{discussions}, from which we will extract all the sentences where the findings are referred to, as explained in step seven.

\begin{figure}
\centering
\includegraphics[width=0.8\textwidth]{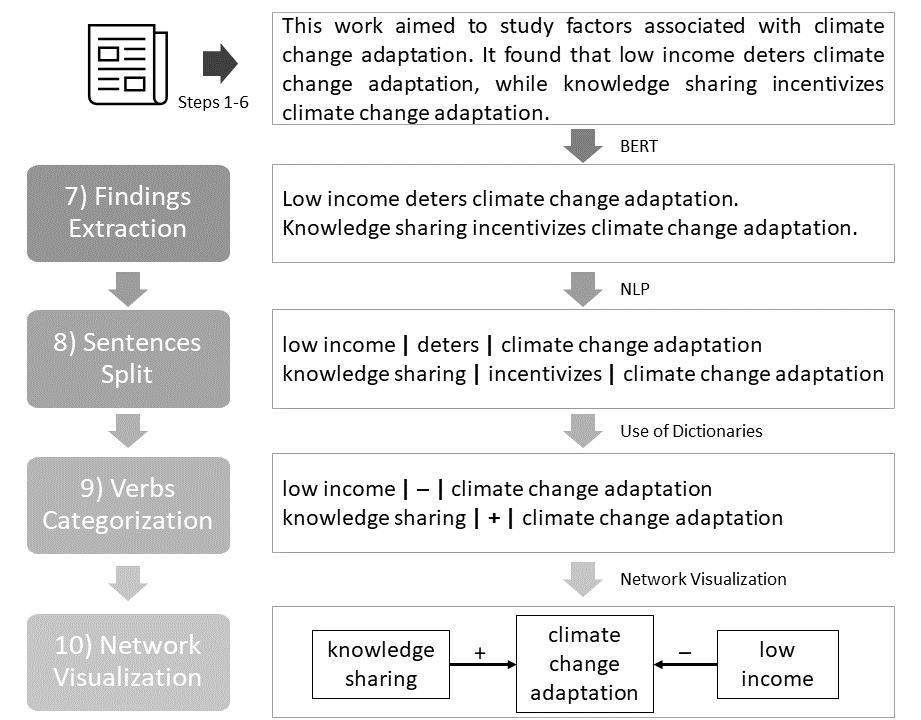}
\caption{Workflow chart diagram} \label{fig2.2}
\end{figure}

\begin{figure}[h]
     \centering
     \begin{subfigure}[b]{0.45\textwidth}
         \centering
         \includegraphics[width=\textwidth]{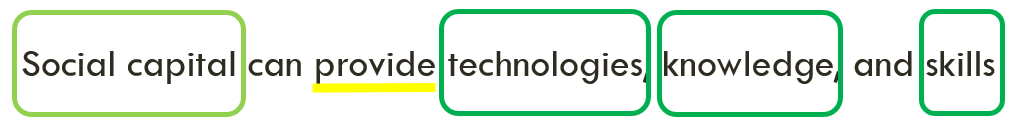}
         \caption{Example sentence of variables relations}
         \label{fig:fig3a}
     \end{subfigure}
     \hfill
     \begin{subfigure}[b]{0.45\textwidth}
         \centering
         \includegraphics[width=\textwidth]{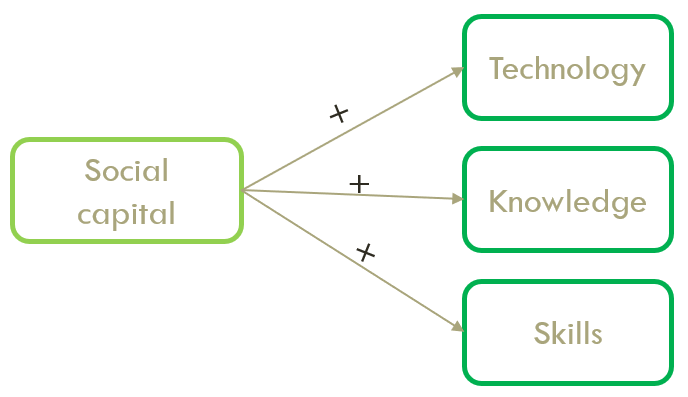}
         \caption{Variables translation into a directed weighted graph.}
         \label{fig:fig3b}
     \end{subfigure}
        \caption{Example of a sentence transformation into directed network.}
        \label{fig:fig3}
\end{figure}

Steps seven to ten comprise the contribution of this work (Fig. \ref{fig2.2}). Step seven corresponds to {\itshape Findings Identification}. For this, we used Bidirectional Encoder Representations from Transformers (BERT) sentence embedding \cite{devlin_bert_2019,devlin_open_2018}, which transforms each sentence into a numerical vector that later could be classified as part of the article’s findings by comparing them against the embedding from three different prompts\footnote{1) What are the factors associated with adaptation?; 2) What are the drivers of adaptation?; and 3) What are the constraints of adaptation?}. From here, we classified a sentence as a finding if it had an average dot-product equal to or bigger than 0.5, implying the sentences are similar. Step eight corresponds to {\itshape Sentences Split}. It detects the subject and objects connected by the verbs (Figure \ref{fig:fig3a}). For this, we used the Parts-Of-Speech returned in step five. Once the sentences are split into three parts (i.e., subject, verb, and object), we homogenize and simplify the information. Homogenizing means that any time we found subjects or objects that conveyed the same by means of permuting the words, e.g. \say{climate change adaptation} or \say{adaptation climate change}, then those sections would be set to the permutation that appeared more often in the findings. Simplifying refers to extracting the subject's and object's words that are more likely to be the central information. For example, after all the steps, a sentence would be \say{well access market extension help african farmer adapt climate change}, this sentence would be split into: subject– \say{well access market extension}; verb– \say{help}; and object– \say{african farmer adapt climate change}. The subject and object central terms would be \say{access market} and \say{adapt climate change}, respectively. Therefore, to extract the central terms, we built a function to find the more likely words to appear and to appear together with other words. For this, we use \texttt{TfidfVectorizer} from the package \texttt{sklearn} \cite{noauthor_scikit-learn_nodate}. Homogenizing and simplifying are important because, otherwise, the network visualization consists of too many different nodes that are difficult to read and interpret.
  
For step nine, {\itshape Verbs Categorization}, we built a dictionary of the verbs commonly used in describing findings. We created a sorted database with all the verbs found in the sentences from step six. Afterward, we read a random sample of ten sentences for each verb to categorize the verb as positive, negative, neutral, dependent, or none. {\itshape Positive} means that the verb denotes a positive relation, e.g., increase, improve, enhance. {\itshape Negative} refers to a verb that denotes a negative association, e.g., reduce, prevent, constrain. {\itshape Neutral} denotes an undirected association, for example, relate, link, associate. {\itshape Dependent} refers to those verbs that require more information to indicate a possible association and direction. For example, \say{have} could denote a positive association if the following words contain the word positive (e.g., have a positive correlation) and a negative association if the following words contain the word negative (e.g., have a negative association). Other verbs in this category are affect, influence, show, etc. However, we assign the verb’s sign as {\itshape positive} or {\itshape negative} once these verbs are identified based on the extra information. Finally, {\itshape none} refers to those verbs that do not denote an association between the words, such as adopt, cope, and implement. Appendix B shows the complete list of verbs categorized as positive, negative, and dependent.
 
Finally, step ten corresponds to {\itshape Network Visualization}. Our implicit assumption is that the findings can be plotted as directed weighted graphs easily interpreted as possible associations depending on the verb. For this, we dropped out the sentences linked by verbs in the None category. Researchers in the social sciences tend to use vocabulary very carefully, as their results may not denote causation but correlation \cite{frankenhuis_strategic_2023}. Therefore, it is advisable to interpret their results as associations instead of causations. For example, Figure \ref{fig:fig3a} can be translated into the graph plotted in Figure \ref{fig:fig3b}. In Figure \ref{fig:fig3b}, the sign corresponds to the relation between the words based on the verb: +: positively associated; +/-: neutrally associated; and -: negatively associated. In the following section, we further elaborate on the network visualization.

\subsection{Aesthetics of the Network Visualization} 

For all the networks, we propose the following aesthetics. First, the words are arranged in concentric circles where the central words denote terms that have the highest node degree. To place the nodes in concentric circles, we only considered whether the word appears in at least one finding sentence of each article. However, we did not add up the frequency it appears in the article (i.e., by article, its maximum value is one). Suppose there are $n$ articles and $p$ words in the database, then the nodes’ degree ($D$) would be given by Equation 1. In Equation 1, $I_{\{.\}}$  denotes the indicator function, which returns 1 when the word $W_j$ appears in the findings sentences of article $A_i$ and 0 in any other case. \\
\begin{equation}
D(W_j)=\sum_{i}I_{\{A_{i}(W_j)\}} \quad \textrm{i=[1,n];  j=[1,p]}
\end{equation}
Second, the edge weight denotes the weighted frequency the words appeared related in the findings. To measure the edge weight, we use the ratio of the number of times the relationship between the words appeared in one article and then added up these ratios for all the articles. The edge weight ($EW$) would be given by Equation 2. In Equation 2, $A_{i,f}(W_k,W_l)$ is 1 when there is a finding sentence ($f$) that associates words $W_k$ and $W_l$ of article $i$ and 0 in any other case.\\
\begin{equation}
\begin{split}
EW(W_k,W_l) & =\frac{\sum_i\frac{\sum_{f}A_{i,f}(W_k,W_l)}{\sum_{f}\sum_{j1}\sum_{j2}A_{i,f}(W_{j1},W_{j2})}}{D(W_k,W_l)}\\
\textrm{with } D(W_k,W_l) & =\sum_{i}I_{\{A_{i}(W_k,W_l)\}}
\end{split}
\end{equation}
where $i=[1,n];  j1,j2=[1,p]; f\in\{A_i\textrm{ findings}\}$.

Third, the edge color shows the most frequent weighted verb sign linking the words. This is measured by the ratio of the number of times the verb sign appeared in the relation between the words in one article divided by the number of times the verb sign appeared between the words in all the articles. The verb sign ($VS$) would be given by Equation 3. In Equation 3, $A_i(s,W_k,W_l)$ is 1 when there is a finding sentence that associates words $W_k$ and $W_l$ by the sign and 0 in any other case.\\
\begin{equation}
EW(s,W_k,W_l)=\frac{\sum_i\frac{\sum_{f}A_{i,f}(s,W_k,W_l)}{\sum_{f}\sum_{j1}\sum_{j2}A_{i,f}(s,W_{j1},W_{j2})}}{D(W_k,W_l)}
\end{equation}
where $i=[1,n];  j1,j2=[1,p]; f\in\{A_i\textrm{ findings}\}$.\\

Finally, the color assigned to the node denotes the cluster to which it belongs. However, the color of the verb sign could be applied to the node, when one node is the source or target of all the other nodes. In this case the visualization would not show the edges. The following section shows how these aesthetics help with the interpretability of the NLP-supported literature review.

\section{Network Visualization of NLP-supported Findings vs. Text Summarization Systems} 

This section aims to show how networks simplify the interpretation of the relationships between the extracted concepts while showing more information than results obtained using text summarization systems. It does so by using the results from Dang's et al. \cite{dang_factors_2019} literature review as the gold standard when comparing the results from our network against the results from text summarization systems. For this, we apply our methodology and the text summarization systems to 91\% of the articles analyzed by Dang et al. \cite{dang_factors_2019} (Table \ref{tab:Tab1}, Appendix A). Specifically, we use the text from the articles' abstracts, analysis, conclusions, and discussion. For the summarization, we also use the sentences from which the network visualization is created, in other words, the sentences use in step eight described in section {\itshape 2.1 General Overview of the Algorithm}. This makes both methodologies comparable.

\subsection{Network Visualization of NLP-supported Findings}
Fig. \ref{fig4} shows the clustered network of nodes that target nodes containing the regular expressions associated with farmers' climate change adaptation. In Fig. \ref{fig4}, the centers of the main clusters are ``adaptation climate change'', ``adaptation measure'', ``adaptation strategy'', ``climate adaptation'', ``farmer adaptation'', ``adaptation climate change'', and ``use adaptation measure''. Those centers reflect the main topics of the clusters. 

Based on the degree (Eq. 1) and edge weights (Eq. 2) definitions, we interpret Fig. \ref{fig4} in the following way. If we focus on the cluster ``adaptation climate change'', we see that the terms that appear more often across articles (among others) are ``effect climate change'' and ``crop''. In contrast, ``non farm income'' and ``soil conservation'' appear less often. The terms that appear more often connected (based on the edge weight) are ``indigenous knowledge financial'' and ``adaptation climate change'', which are positively associated. 

\begin{figure}[!htb]
\centering
\includegraphics[width=\textwidth]{CCA_clust2.png} % , angle=90
\caption{Clustered networks of nodes that target nodes containing the word ``adaptation''.} \label{fig4}
\end{figure}

However, as we aim to show and describe all the factors associated with farmers' adaptation to climate change, we have decided to uncluster the network and rename all the targeted nodes as ``climate change adaptation'' (Fig. \ref{fig5}). This way, the color of the nodes reflects the edge color (i.e., the color of the most common association: positive ``+'', neutral ``+/-'', or negative ``-'') between the nodes and the central term (``climate change adaptation''). This eases the reading of the results. 

In the next paragraphs, we describe the nodes that, according to our knowledge, can be considered as socio-economic factors and those nodes that resemble climate change motivators according to other fields (such as psychology) \cite{berrang-ford_systematic_2021}. As Fig. \ref{fig5} shows, the factors that are positively associated with farmers' climate change adaptation are, in descending order relative to their degree: climate change, extension service, education, access credit, information climate change, government policy, age household head, farming experience, effect climate change, risk, social capital, adaptive capacity community, agricultural extension service, family size, knowledge, access technology, gender household head, household size, farmer analytical capacity, labor force, migration labor, local state institution, involvement state institution, social institution, access extension, soil conservation, context, network, access market, fertilizer subsidy scheme, soil fertility, livestock size, age, information, technology, household size, young respondent, knowledge climate change, land ownership, gender, indigenous knowledge financial, farm income, perception climate change, and decision making process.

\begin{comment}
 if i would group those by meaning/category, e.g.  based on dif adaptation constraints:

\textit{`economic' adaptation constraints:} household size, family size, labor force, migration labor, access market, fertilizer subsidy scheme, livestock size, land ownership,  farm income, 

 \textit{`social' adaptation constraints}:  extension service, age household head, farming experience, 
 
 \textit{`human capacity' adaptation constraint}: education; age, household size, young respondent, social capital, adaptive capacity community, agricultural extension service, knowledge,  gender household head, farmer analytical capacity, network,  gender, indigenous knowledge financial, 

 \textit{`governance/institutions/policy' adaptation constraint}: government policy,  local state institution, involvement state institution, social institution, access extension, context, 

 \textit{`financial' adaptation constraints}: access credit, 

 \textit{`information/awareness/technology' adaptation constraint}: information climate change,  access technology, information, technology, knowledge climate change, perception climate change, and decision making process.

 \textit{`bio-physical' adaptation constraint:} climate change, effect climate change, risk, soil conservation, soil fertility,

\end{comment}

The factors that are neutrally associated with farmers' climate change adaptation are policy, cooperation farm level, adaptive capacity, institutional structure, effect climate, farm level practice, technological development, weather, food security, ownership, and government. Finally, the negatively associated factors are distance to market, credit, forecasting, migration household, employment, and government support.

\begin{figure}[!htb]
\centering
\includegraphics[width=\textwidth]{CCA_2.png} % , angle=90
\caption{The complete network of all terms associated with farmers' climate change adaptation in the processed literature.} \label{fig5}
\end{figure}

\subsection{Text Summarization Systems}

Performing Literature Reviews using text summarization methods have two difficulties. First, text summarization methods perform well on short texts, such as blog posts and news \cite{koh_empirical_2023}, but not in long texts, such as articles. It is more difficult to summarise long text because readers tend to be interested on different sections of the work \cite{koh_empirical_2023}. This means that researchers need to train the models based on their own necessities. Therefore, in this section, we used the sentences from which the network visualization was created, i.e. we focus on the articles' findings.

The second difficulty stems from the amount of articles that need to be summarise. For this, researchers have proposed some methodologies to summarize text that comes from more than one source. These are commonly called Multi-Document Summarization \cite{wang_integrating_2011,cao_improving_2017,cao_ranking_2015}. These methodologies normally rely on clustering the documents to later summarize each cluster using either extractive or abstractive summarization methods \cite{wang_integrating_2011,cao_improving_2017,cao_ranking_2015}. However, in our case, we decided to cluster the sentences instead of the documents and then summarize those clusters. This is a better strategy when summarising content for the literature reviews, as the sentences may refer to different findings (e.g., how age associates with farmers' climate change adaptation).

To perform the text summarization, we use extractive, abstractive, and hybrid methods on the 43 articles' findings sentences that were considered when creating the Network Visualization of NLP-supported Findings. The sentences were clustered into 18 groups using the Non-Negative Matrix Factorization from the SciKit-Learn package \cite{scikit_learn_non-negative_nodate}. We decided on the number of clusters and clusterization method by reading the sentences in those clusters and checking how cohesive the groups were. To summarize the sentences, we used the Python packages \texttt{sumy} \cite{belica_sumy_nodate} and \texttt{transformers} \cite{hugging_face__nodate}. 

The summarization methods used are the following. For the text extractive summarization, we used the Text Rank and Lex Rank methods \cite{giarelis_abstractive_2023}, as they are among the most promising and efficient extractive summarization methods \cite{dsilva_impact_2023}. For this, we constrained the output to at most 2 sentences per cluster (the summaries are available in Appendix C). 

For the abstractive summarization, we decided to use the Bidirectional Auto-Regressive Transformers (BART) and the Pretraining with Extracted Gap-Sentences for Abstractive Summarization (PEGASUS) methodologies, as they are the ones reported to achieve the best results \cite{giarelis_abstractive_2023}. For each method, we used either the full cluster's sentences or the most relevant sentences according to the aforementioned extractive methodologies. The latter was necessary because the abstractive summarization systems have an upper bound on the length of text passed to summarize \cite{hugging_face__nodate}. So, whenever the text length was longer than the upper bound, we used the extractive system to choose 75\% of the most important sentences in the articles' findings (this was necessary to force the algorithm to rank the sentences). Then, we concatenated the number of sentences necessary so that the upper bound was not reached for each of the abstractive summarization systems (the summaries are available in Appendix C).

As a quantitative measure of the similarity of these summaries and Dang's et al. \cite{dang_factors_2019} findings, we use the mean of the different ROUGE measures (Table \ref{tab:Tab2}). The different ROUGE measures evaluate the quality of the summaries by counting the unit overlap between the summaries and the reference summary \cite{lin_rouge_2004}. Table \ref{tab:Tab2} shows that, on average, the methods BART-TextRank and TextRank have similar performance when compared to Dang's et al. \cite{dang_factors_2019} findings, while BART-LexRank differs by one percentage point. Whilst, LexRank, pegasus-TextRank, and pegasus-LexRank performed worse.

\begin{table}[ht]
\centering
\caption{Rouge metrics and their mean for the different summarization systems.} \label{tab:Tab2}
\begin{tabular}{rlrrrrr}
  \hline
 & \multicolumn{1}{c}{Summarization} & \multirow{2}{*}{rouge1} & \multirow{2}{*}{rouge2} & \multirow{2}{*}{rougeL} & \multirow{2}{*}{rougeLsum} & \multirow{2}{*}{mean} \\ 
 & \multicolumn{1}{c}{Method} &  &  &  &  &  \\ 
  \hline
  1 & LexRank & 0.38 & 0.11 & 0.15 & 0.15 & 0.20 \\ 
  2 & TextRank & 0.44 & 0.12 & 0.17 & 0.17 & 0.23 \\ 
  3 & BART-LexRank & 0.45 & 0.12 & 0.16 & 0.16 & 0.22 \\ 
  4 & BART-TextRank & 0.45 & 0.13 & 0.17 & 0.17 & 0.23 \\ 
  5 & pegasus-LexRank & 0.35 & 0.12 & 0.15 & 0.15 & 0.19 \\ 
  6 & pegasus-TextRank & 0.35 & 0.12 & 0.15 & 0.15 & 0.19 \\ 
   \hline
\end{tabular}
\end{table}

\subsection{Human Comparison of Both Methodologies}

In this section, we analyze whether the Network Visualization of NLP-supported Findings better captures farmers' factors associated with climate change adaptation compared to Text Summarization Systems. To do so, we rely on two metrics applied to the network and the summarization systems: 1) their quality to summarise the articles' findings; and 2) their ability to capture the same results as Dang's et al. \cite{dang_factors_2019}. For this, we asked six PhD students, that have knowledge on the climate change literature, to read Dang's et al. \cite{dang_factors_2019} findings and then grade the quality of the network and the summaries (Table 2) and their ability to capture the same findings as Dang's et al. \cite{dang_factors_2019} (Table 3). For the former, the students were asked to give a score between 0 to 10, where 0 represents not understandable and 10 perfectly understandable. For the latter, the students marked whether Dang's et al. \cite{dang_factors_2019} findings were present in the network and summaries (Table 3).

Table 2 shows the mean and variance of the students' scores. In Table 2, the Network Visualization had the highest average score and the smaller variance. This means that compared to the summarization results, the Network Visualization is easier to read. However, as the Network Visualization scored 6.7 (when the maximum is 10), this also means that the Network Visualization needs to be improved if we want to ease its readability.

\begin{table}[ht]
\centering
\caption{Quality of the results according to students' scores.} \label{tab:Tab3}
\begin{tabular}{r|l|c|c}
  \hline
 \multicolumn{2}{c|}{Method} & Mean & Variance \\ 
  \hline
  \multicolumn{2}{c|}{Network Visualization}  & 6.7 & 0.2\\ 
  \hline
\parbox[t]{4mm}{\multirow{6}{*}{\rotatebox[origin=c]{90}{Summarization}}} & LexRank & 5.4 & 3.8\\ 
   & TextRank & 3.6 & 2.3 \\ 
   & BART-LexRank & 5.4 & 4.3 \\ 
   & BART-TextRank & 4.6 & 2.8 \\ 
   & pegasus-TextRank & 4.2 & 1.7 \\ 
   & pegasus-LexRank & 5.4 & 3.3 \\ 
   \hline
\end{tabular}
\end{table}

Table 3 shows whether the factors included in Dang's et al. \cite{dang_factors_2019} literature review are connected in the Network Visualization of NLP-supported Findings and whether they are mentioned in the Text Summarization Systems based on the students results. Table 3 also shows whether the direction of the association (Sign) is the same as the one found by Dang et al. \cite{dang_factors_2019}. From the 30 factors associated with farmers' adaptation to climate change, our methodology captured (19/30) 61\% of the factors, from which 79\% had the correct association with farmers' adaptation to climate change. In the case of the summarization systems, on average, they captured 46\% of the factors with an average of 76\% correct associations. Based on these results, using our methodology, we were better able to come up with similar conclusions as Dang et al. \cite{dang_factors_2019} compared to summarization systems.

\begin{adjustbox}{angle=90,scale=0.75,caption={Comparison of results obtained using the Network Visualization of NLP-supported Findings and the Text Summarization.},float=table}\label{tab:Tab4}
\centering
\begin{tabular}{|rr|c|cc|cc|cc|cc|cc|cc|cc|}
  \hline
 & \multicolumn{2}{c|}{Literature} & \multicolumn{2}{|c|}{\multirow{2}{*}{Network}}  & \multicolumn{2}{|c|}{LexRank} & \multicolumn{2}{|c|}{TextRank} &  \multicolumn{2}{|c|}{\multirow{2}{*}{BART-LR}} & \multicolumn{2}{|c|}{\multirow{2}{*}{BART-TR}} & \multicolumn{2}{|c|}{\multirow{2}{*}{PEGASUS-LR}} & \multicolumn{2}{|c|}{\multirow{2}{*}{PEGASUS-TR}}\\ 
 & \multicolumn{2}{c|}{Review} & \multicolumn{2}{|c|}{}  & \multicolumn{2}{|c|}{(LR)} & \multicolumn{2}{|c|}{(TR)} &  \multicolumn{2}{|c|}{} & \multicolumn{2}{|c|}{} & \multicolumn{2}{|c|}{} & \multicolumn{2}{|c|}{}\\ 
 \hline
 & Factor & Association & Factor & Sign & Factor & Sign & Factor & Sign & Factor & Sign & Factor & Sign & Factor & Sign & Factor & Sign \\ 
  \hline
  1 & Age & +/- & \cmark & \cmark & \cmark & \cmark & \cmark & \cmark & \cmark & \cmark & \cmark & \cmark & \cmark & \cmark & \cmark & \cmark \\ 
  2 & Young Age & + &  &  &  &  &  &  &  &  &  &  &  &  &  &  \\ 
  3 & Old Age & - &  &  &  &  &  &  &  &  &  &  &  &  &  &  \\ 
  4 & Gender & +/- & \cmark &  &  &  &  &  &  &  &  &  &  &  &  &  \\ 
  5 & Male & + &  &  &  &  &  &  &  &  &  &  &  &  &  &  \\ 
  6 & Female & - &  &  &  &  &  &  &  &  &  &  &  &  &  &  \\ 
  7 & Education & + & \cmark & \cmark & \cmark & \cmark & \cmark & \cmark & \cmark & \cmark & \cmark & \cmark & \cmark & \cmark & \cmark & \cmark \\ 
  8 & Household Size & + & \cmark & \cmark & \cmark & \cmark & \cmark & \cmark & \cmark & \cmark & \cmark & \cmark & \cmark & \cmark & \cmark & \cmark \\ 
  9 & Income & + & \cmark & \cmark &  &  &  &  & \cmark & \cmark & \cmark & \cmark & \cmark & \cmark & \cmark & \cmark \\ 
  10 & Agricultural Extension & + & \cmark & \cmark & \cmark & \cmark & \cmark & \cmark & \cmark & \cmark & \cmark & \cmark & \cmark & \cmark & \cmark & \cmark \\ 
  11 & Credit & + & \cmark & \cmark & \cmark & \cmark & \cmark & \cmark & \cmark & \cmark & \cmark & \cmark & \cmark & \cmark & \cmark & \cmark \\ 
  12 & Information & + & \cmark & \cmark & \cmark & \cmark & \cmark & \cmark & \cmark & \cmark & \cmark & \cmark & \cmark & \cmark & \cmark & \cmark \\ 
  13 & Information Climate Change & - & \cmark & \cmark & \cmark & \cmark & \cmark &  & \cmark &  & \cmark &  & \cmark &  & \cmark & \cmark \\ 
  14 & Information Climate & + & \cmark & \cmark & \cmark & \cmark & \cmark & \cmark & \cmark & \cmark & \cmark & \cmark & \cmark &  & \cmark & \cmark \\ 
  15 & Technology & + & \cmark & \cmark & \cmark & \cmark & \cmark & \cmark & \cmark & \cmark & \cmark & \cmark & \cmark & \cmark & \cmark & \cmark \\ 
  16 & Institutional Arrangements on Land & + &  &  & \cmark & \cmark &  &  & \cmark & \cmark & \cmark & \cmark &  &  &  &  \\ 
  17 & Non-Governmental Organizations & + &  &  &  &  &  &  & \cmark & \cmark &  &  &  &  &  &  \\ 
  18 & State Institutions & +/- & \cmark & \cmark & \cmark & \cmark & \cmark &  & \cmark &  & \cmark &  &  &  &  &  \\ 
  19 & Government & - & \cmark & \cmark & \cmark &  & \cmark &  & \cmark &  & \cmark &  & \cmark &  & \cmark & \cmark \\ 
  20 & Norms & - & \cmark &  &  &  &  &  &  &  &  &  &  &  &  &  \\ 
  21 & Socio-Cultural & - & \cmark &  &  &  &  &  &  &  & \cmark &  &  &  & \cmark & \cmark \\ 
  22 & Climate Change Concern & + & \cmark &  & \cmark &  &  &  &  &  & \cmark & \cmark &  &  &  &  \\ 
  23 & Climate Change Denial & - &  &  &  &  &  &  &  &  &  &  &  &  &  &  \\ 
  24 & Perceived Risk & +/- &  &  & \cmark &  & \cmark &  & \cmark &  & \cmark & \cmark & \cmark & \cmark & \cmark & \cmark \\ 
  25 & Perceived Effectiveness & +/- & \cmark & \cmark &  &  & \cmark &  &  &  &  &  & \cmark & \cmark & \cmark &  \\ 
  26 & Maladaptation & +/- &  &  &  &  &  &  &  &  &  &  &  &  &  &  \\ 
  27 & Perceived Peer Pressure & +/- &  &  &  &  &  &  &  &  &  &  &  &  &  &  \\ 
  28 & Perceived Prices Increase  & +/- &  &  &  &  &  &  &  &  &  &  &  &  &  &  \\ 
  29 & Human Cognition & + & \cmark & \cmark &  &  &  &  &  &  &  &  &  &  &  &  \\ 
  30 & Scepticism & - & \cmark & \cmark &  &  &  &  &  &  &  &  &  &  &  &  \\ 
  \hline
   & \multirow{2}{*}{Correct} & Total & 21 & 14 & 12 & 11 & 12 & 10 & 13 & 12 & 14 & 13 & 12 & 9 & 11 & 9 \\ 
   &  & \% & 61\% & 79\%* & 45\% & 79\%* & 42\% & 62\%* & 48\% & 73\%* & 52\% & 75\%* & 42\% & 77\%* & 45\% & 93\%* \\ 
   \hline
   \multicolumn{17}{r}{\raggedright * Percentage of factors with correct sign from those found in the results. \par}
\end{tabular}
\end{adjustbox}

\section{Conclusions}

This article proposed an interpretable NLP algorithm and descriptive visualizations to summarize articles' findings when performing systematic literature reviews. Our approach aimed to avoid using Black-Box algorithms and opaque machine learning methodologies to extract and synthesize information. On the one hand, our method shows that the use of tailored verb dictionaries helps to interpret the results shown in the networks. On the other hand, sensible visualizations of the results also facilitate synthesizing the information. Furthermore, according to study participatants, our methodology performs better than extractive, abstractive, and hybrid summarization methods.

Our methodology, Network Visualization of NLP-supported Findings, was able to outperform summarization methods because it eases the identification and visualization of the variables that are more often connected in the articles. This is because of their network location: the closer to the center, the bigger the node degree. For example, in Figure \ref{fig5}, factors like access to credit, extension service, and education are more likely to be referred to as farmers' adaptation factors in the articles. In contrast, perception climate change, indigenous knowledge, and gender appear less often. So, the visualization gives not only the direction and the frequency of the association between the words but also the frequency the word appears in the articles relative to others, which is something that is lost when using summarization methods.

However, results derived from this methodology are intended to help researchers synthesize information from large amounts of articles. Even though results interpretation should be made cautiously and always based on previous knowledge, this methodology can help researchers keep track of the findings from their areas of expertise.

\section{Limitations and Future Work}

Our work has some limitations that we would like to acknowledge. Compared to machine learning, our methodology requires that researchers subjectively classify the verbs found in sentence samples as conveying positive, negative, or neutral associations between words. However, we do not believe this to be a problem as long as researchers openly share the dictionary they used when implementing this algorithm. Furthermore, as the results are descriptive, everything can be tracked down to check why some words show specific associations and to which article the relations belong. Another limitation is that we based our network visualizations on categorical definitions of the signs: the sign that appeared most frequently. This definition has the limitation of not considering that some variables might have very close percentages of associated positive, negative, and neutral verbs. Future work can explore the possibility of showing signs' colors in terms of continuous variables.

\textbf{Funding:} This work was supported by the Netherlands Organization for Scientific Research NWO VIDI grant number 191015. 

\textbf{Code Availability:} The codes to replicate this work will be available upon publication of the article at: (\url{https://github.com/SC3-TUD}) upon publication of the article.

\bibliographystyle{splncs04}
\bibliography{GilClavel_NLP_and_Networks_for_LR}

\newpage

\appendix

\section{Articles used to build the database}

\begin{table}[ht]
\caption{List of articles used in Dang et al. \cite{dang_factors_2019} and that form our text database.} \label{tab:Tab1}
\centering
\resizebox{0.8\textwidth}{!}{
\begin{tabular}{rlrl}
  \hline
 & Reference & & Reference \\ 
  \hline
1 & Abid et al. \cite{abid_climate_2016} & 23 & Hisali et al. \cite{hisali_adaptation_2011} \\ 
  2 & Adger \cite{neil_adger_social_1999} & 24 & Idrisa et al. \cite{idrisa_analysis_2012} \\ 
  3 & Alam \cite{alam_farmers_2015} & 25 & Jones and Boyd \cite{jones_exploring_2011} \\ 
  4 & Antwi-Agyei et al. \cite{antwi-agyei_barriers_2015} & 26 & Komba and Muchwaponda \cite{komba_adaptation_2012} \\ 
  5 & Anyoha et al. \cite{anyoha_socio-economic_2013} & 27 & Kuehne \cite{kuehne_how_2014} \\ 
  6 & Arbuckle et al. \cite{arbuckle_farmer_2013} & 28 & Mandleni and Anim \cite{mandleni_climate_2011} \\ 
  7 & Arimi \cite{arimi_determinants_2014} & 29 & Muller and Shackleton \cite{muller_perceptions_2014} \\ 
  8 & Balew et al. \cite{balew_determinants_2014} & 30 & Ndambiri et al. \cite{ndambiri_assessment_2012} \\ 
  9 & Baudoin \cite{baudoin_enhancing_2014} & 31 & Nielsen and Reenberg \cite{nielsen_cultural_2010} \\ 
  10 & Below et al. \cite{below_can_2012} & 32 & Oyekale and Oladele \cite{oyekale_determinants_2012} \\ 
  11 & Biggs et al. \cite{biggs_agricultural_2013} & 33 & Ozor and Cynthia \cite{ozor_difficulties_2011} \\ 
  12 & Bryan et al. \cite{bryan_adaptation_2009} & 34 & Pauw \cite{pauw_role_2013} \\ 
  13 & Comoé and Siegrist \cite{comoe_relevant_2015} & 35 & Salau et al. \cite{salau_knowledge_2013} \\ 
  14 & Dang et al. \cite{dang_understanding_2014} & 36 & Sarker et al. \cite{sarker_assessing_2013} \\ 
  15 & De Jalón et al. \cite{garcia_de_jalon_behavioural_2015} & 37 & Shiferaw \cite{shiferaw_smallholder_2014} \\ 
  16 & Deressa et al. \cite{deressa_determinants_2009} & 38 & Sofoluwe et al. \cite{sofoluwe_farmers_2011} \\ 
  17 & Deressa et al. \cite{deressa_perception_2011} & 39 & Truelove et al. \cite{truelove_socio-psychological_2015} \\ 
  18 & Esham and Garforth \cite{esham_agricultural_2013} & 40 & Tucker et al. \cite{tucker_perceptions_2010} \\ 
  19 & Fosu-Mensah et al. \cite{fosu-mensah_farmers_2012} & 41 & Wilk et al. \cite{wilk_adaptation_2013} \\ 
  20 & Gbetibouo et al. \cite{gbetibouo_modelling_2010} & 42 & Yegbemey et al. \cite{yegbemey_farmers_2013} \\ 
  21 & Gebrehiwot and van der Veen \cite{gebrehiwot_farm_2013} & 43 & Yila and Resurreccion \cite{othniel_yila_determinants_2013} \\ 
  22 & Hassan and Nhemachena \cite{hassan_determinants_2008} &  &  \\ 
   \hline
\end{tabular}}
\end{table}

\newpage

\section{Verbs Findings Dictionary}

In this section, we present the list of verbs used in the dictionary. In the lists, the dependent verbs were already listed under the categories positive or negative.

\begin{itemize}
\item List of verbs denoting a positive association:
\begin{multicols}{4}
affect positive\\
accelerate\\
access\\
achieve\\
adapt\\
alleviate\\
augment\\
avail\\
be gain\\
be promote\\
be protect\\
be support\\
bring\\
build\\
can transfer\\
combat\\
compensate\\
contribute\\
create\\
determine\\
drive\\
enable\\
enact\\
encourage\\
enhance\\
enrich\\
explain\\
facilitate\\
foster\\
fund\\
gain\\
galvanize\\
generate\\
have increase\\
have positive\\
help\\
improve\\
increase\\
induce\\
influence positive\\
inform\\
intensify\\
introduce\\
join\\
opt\\
perceive positive\\
plant\\
predict\\
prepare\\
promote\\
provide\\
raise\\
rear\\
rehabilitate\\
reimburse\\
require\\
resolve\\
restore\\
scale\\
should include\\
show positive\\
speed\\
strengthen\\
support\\
tackle\\
tap\\
trust\\
use positive\\
upgrade\\
\end{multicols}

\item List of verbs denoting a neutral association:
\begin{multicols}{4}
account\\
act\\
address\\
adjust\\
appear\\
apply\\
associate\\
assume\\
attribute\\
be\\
be determine\\
be relate\\
be report\\
be root\\
change\\
combine\\
connect\\
control\\
correlate\\
deem\\
diversify\\
dominate\\
emerge\\
engage\\
evidence\\
evolve\\
impact\\
intend\\
involve\\
keep\\
know\\
link\\
maintain\\
manage\\
mediate\\
mention\\
need\\
neutralize\\
offset\\
possess\\
reach\\
reflect\\
relate\\
represent\\
signify\\
stem\\
understand\\
\end{multicols}

\item List of verbs denoting a negative association:
\begin{multicols}{4}
affect negative\\
constrain\\
damage\\
decline\\
decompose\\
decrease\\
erode\\
exacerbate\\
fail\\
have negative\\
have prevent\\
hinder\\
ignore\\
influence negative\\
limit\\
minimize\\
mitigate\\
perceive negative\\
prevent\\
reduce\\
restrict\\
retard\\
show negative\\
tradeoff\\
use negative\\
weaken\\
\end{multicols}

\end{itemize}

\newpage

\end{document}